%% file: main.tex
\documentclass[10pt,twocolumn,letterpaper]{article}
\PassOptionsToPackage{table,dvipsnames}{xcolor}
\usepackage[pagenumbers]{cvpr} 

\input{preamble}

\definecolor{cvprblue}{rgb}{0.21,0.49,0.74}
\usepackage[pagebackref,breaklinks,colorlinks,allcolors=cvprblue]{hyperref}

\def\paperID{1494} 
\def\confName{CVPR}
\def\confYear{2025}

\title{Show and Segment: Universal Medical Image Segmentation via In-Context Learning}

\author{Yunhe Gao$^{2}$, Di Liu$^{2}$, Zhuowei Li$^{2}$, Yunsheng Li$^{1}$, Dongdong Chen$^{1}$, Mu Zhou$^{2}$, Dimitris N. Metaxas$^{2}$\\\\
$^1$Microsoft GenAI~~~$^2$Rutgers University\vspace{-0em}
}

\begin{document}
\maketitle
\input{sec/0_abstract}    
\input{sec/1_intro}
\input{sec/2_related_work}
\input{sec/3_method}

\input{sec/4_experiment}

\input{sec/5_discussion}

{
    \small
    \bibliographystyle{ieeenat_fullname}
    \bibliography{reference.bib}
}

\input{sec/X_suppl}

\end{document}

%% file: preamble.tex
\definecolor{lightgray}{rgb}{0.92, 0.92, 0.92}

\usepackage[utf8]{inputenc} 
\usepackage[T1]{fontenc}    
\usepackage{url}            
\usepackage{booktabs}       
\usepackage{amsfonts}       
\usepackage{nicefrac}       
\usepackage{microtype}      
\usepackage{amsmath}
\usepackage{amssymb}
\usepackage{graphicx}
\usepackage{algorithm}
\usepackage{algpseudocode}
\usepackage{enumitem}
\usepackage{multirow}
\usepackage{wrapfig}
\usepackage{epstopdf}

%% file: sec/0_abstract.tex
\begin{abstract}
Medical image segmentation remains challenging due to the vast diversity of anatomical structures, imaging modalities, and segmentation tasks. While deep learning has made significant advances, current approaches struggle to generalize as they require task-specific training or fine-tuning on unseen classes. We present \textbf{Iris}, a novel \textbf{I}n-context \textbf{R}eference \textbf{I}mage guided \textbf{S}egmentation framework that enables flexible adaptation to novel tasks through the use of reference examples without fine-tuning. At its core, Iris features a lightweight context task encoding module that distills task-specific information from reference context image-label pairs. This rich context embedding information is used to guide the segmentation of target objects. By decoupling task encoding from inference, Iris supports diverse strategies from one-shot inference and context example ensemble to object-level context example retrieval and in-context tuning. Through comprehensive evaluation across twelve datasets, we demonstrate that Iris performs strongly compared to task-specific models on in-distribution tasks. On seven held-out datasets, Iris shows superior generalization to out-of-distribution data and unseen classes. Further, Iris's task encoding module can automatically discover anatomical relationships across datasets and modalities, offering insights into medical objects without explicit anatomical supervision.

\end{abstract}

%% file: sec/1_intro.tex
\section{Introduction}
\label{sec:intro}

The accurate segmentation of anatomical structures in medical images is fundamental for clinical practice and biomedical research, enabling precise diagnosis~\cite{de2018clinically,shen2015multi} and treatment planning~\cite{nestle2005comparison}. While deep learning has demonstrated remarkable success~\cite{liu2021review,wang2022medical}, the vast diversity of anatomical structures, imaging modalities, and clinical tasks poses long-standing challenges for developing truly generalizable solutions. Current efforts typically focus on disease-specific tasks or a limited set of anatomical structures~\cite{isensee2021nnu,liu2021refined,liu2022transfusion,chang2022deeprecon,gao2022data,gao2024training,he2023dealing,zhangli2022region,liu2021label}, struggling to handle the heterogeneous landscape of medical imaging that spans diverse modalities, body regions, and diseases~\cite{yoon2023domain,niu2024survey}.

\begin{figure}[t]
\begin{center}
\includegraphics[width=0.47\textwidth]{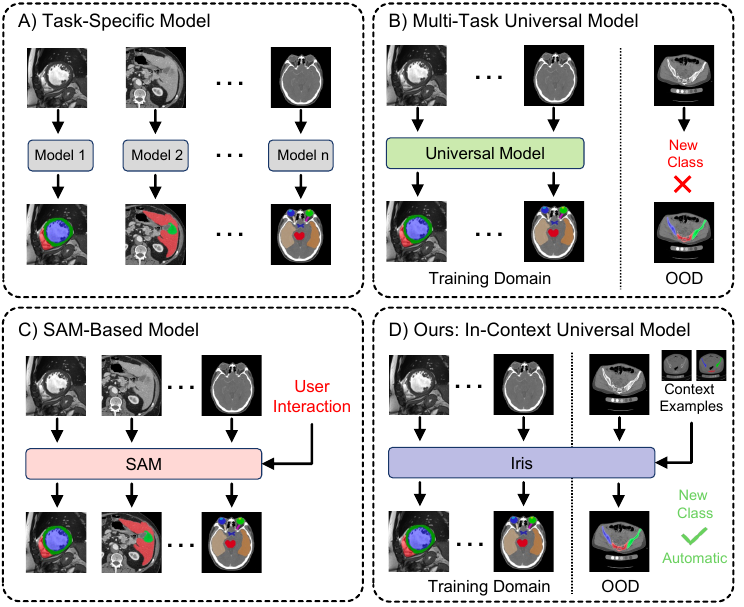}
\end{center}
\vspace{-1em}
\caption{Comparison of medical image segmentation approaches. A) Task-specific models require training separate models for each task, limiting their flexibility and scalability. B) Multi-task universal models can handle diverse tasks and imaging modalities, but fail on novel classes. C) SAM-based foundation models enable flexible segmentation through user interactions, but impractical for high-throughput automated processing. D) Our proposed Iris combines automatic processing with flexible adaptation via in-context learning, enabling both seen and unseen task segmentation without any manual interaction or retraining.}
\label{fig:fig1}
\vspace{-1em}
\end{figure}

These task-specific methods show critical limitations compared to human experts' capabilities. First, existing models often perform poorly on out-of-distribution examples~\cite{zhou2022domain}—a common scenario in medical imaging where variations arise from different imaging centers, patient populations, and acquisition protocols. Second, traditional segmentation models, while achieving a high accuracy on their trained tasks, lack the adaptability to handle novel classes without extensive retraining or fine-tuning~\cite{zhang2023continual}. This dilemma fundamentally limits task-specific models' applicability in dynamic clinical settings and research environments, where new segmentation tasks continue to emerge over the course of real-world practice.

Recent research has explored several directions to address these challenges (Figure \ref{fig:fig1}). Universal medical segmentation models~\cite{zhang2021dodnet,liu2023clip,ye2023uniseg,ulrich2023multitalent} attempt to leverage synergy among multiple tasks across diverse datasets to learn robust representations, yet struggling with unseen classes and requiring fine-tuning. Foundation models with interactive capabilities, such as SAM~\cite{kirillov2023segment} and its medical variants~\cite{zhang2024data,ma2024segment,cheng2023sam,wang2024sam}, offer flexibility via user prompts. But they require multiple interactions for optimal segmentation results, especially for complex 3D structures, and lack the efficiency for large-scale automated analysis. In addition, in-context learning (ICL) methods~\cite{butoi2023universeg,rakic2024tyche} show promise in automatically handling arbitrary new tasks through a few reference examples, but current methods exhibit suboptimal performance compared to task-specific models and suffer from computational inefficiencies, requiring expensive reference encoding during each inference step.

To address these fundamental challenges, we present Iris framework for universal medical image segmentation via in-context learning. At its core, Iris features a lightweight task encoding module that efficiently distills task-specific information from reference image-label pairs into compact task embeddings, which then guide the segmentation of target objects. Unlike existing ICL methods~\cite{butoi2023universeg,rakic2024tyche}, Iris decouples the task definition from query image inference, eliminating redundant context encoding while enabling flexible inference strategies, all coming with high computational efficiency. 

Our main contributions include:
\begin{itemize}
\item A novel in-context learning framework for 3D medical images, enabling a strong adaptation to arbitrary new segmentation tasks without model retraining or fine-tuning.
\item A lightweight task encoding module that captures task-specific information from reference examples, handling medical objects of varying sizes and shapes.
\item Multiple flexible inference strategies suitable for different practical scenarios, including one-shot inference, context ensemble, object-level context retrieval, and in-context tuning.
\item Comprehensive experiments on 19 datasets demonstrate Iris's superior performance across both in-distribution and challenging scenarios, particularly on held-out domains and novel anatomical structures. It extends to reveal the capability of automatically discovering meaningful anatomical relationships across datasets and modalities.
\end{itemize}


%% file: sec/2_related_work.tex
\section{Related Work}
\label{sec:related_work}

\noindent\textbf{Medical Universal Models.}
Universal medical image segmentation models aim to address the data heterogeneity across tasks and modalities while learning generalizable feature representations. Early works include multi-dataset learning through label conditioning~\cite{dmitriev2019learning}, organ size constraints~\cite{zhou2019prior}, and pseudo-label co-training~\cite{huang2020multi}. Recent works are placed on sophisticated task encoding strategies. DoDNet~\cite{zhang2021dodnet} pioneered one-hot task vectors with an extension into TransDoDNet~\cite{xie2023learning} using transformer backbones. Advances include CLIP-driven models using semantic encodings~\cite{liu2023clip}, task-specific heads in MultiTalent~\cite{ulrich2023multitalent}, and modality priors in Hermes~\cite{gao2024training}. UniSeg~\cite{ye2023uniseg} introduced learnable task prompts and MedUniseg~\cite{ye2024meduniseg} unified 2D/3D image handling. Despite of substantial efforts, these universal models all require fine-tuning when assessing unseen classes. In contrast, Iris enables the segmentation of unseen classes only through a single reference image-label pair without any model finetuning.

\noindent\textbf{SAM-based Interactive Models.}
Segment Anything Model (SAM)~\cite{kirillov2023segment} emerges as a shifting paradigm of interactive segmentation via its prompt-based architecture and large-scale training. SAM's success has inspired a range of medical variants. Major examples include  MedSAM~\cite{ma2024segment} with 1.5M image-mask pairs for 2D segmentation, SAM-Med2D~\cite{cheng2023sam} trained on 4.6M images, and SAM-Med3D~\cite{wang2024sam} extending to volumetric data with 22K 3D images. These efforts all require multiple prompts and interactive refinements, especially for analyzing complex objects in 3D scenarios. This interaction-dependent design becomes a bottleneck in high-throughput scenarios that an automated processing of large-scale datasets is much desired. As comparison, Iris addresses this limitation by defining tasks through context pairs, enabling fully automatic segmentation while maintaining a strong adaptability to new tasks.

\begin{figure*}[t]
\begin{center}
\includegraphics[width=0.9\textwidth]{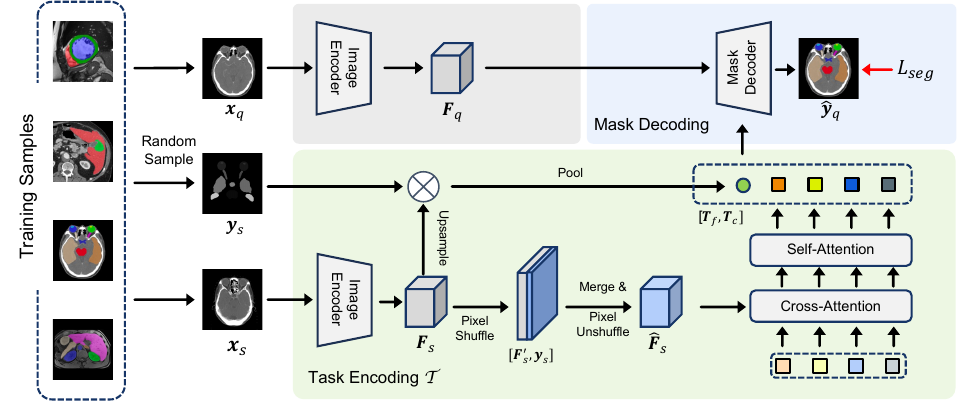}
\end{center}
\vspace{-1em}
\caption{Overview of Iris framework. We design a task encoding module to extract compact task embeddings from reference examples to guide query image segmentation with the mask decoding module, enabling efficient and flexible adaptation to new tasks without finetuning.}
\label{fig:framework}
\vspace{-1em}
\end{figure*}

\noindent\textbf{Visual In-context Learning.} 
In-context learning as introduced by GPT-3~\cite{brown2020language} enables models to handle novel tasks through example-guided inference without a heavy retraining. In the vision community, Painter~\cite{wang2023images} and SegGPT~\cite{wang2023seggpt} pioneered in-context segmentation through a mask image modeling framework. Alternative methods~\cite{liu2023matcher,zhang2023personalize,sun2024vrp} explored SAM-based approaches through cross-image correspondence prompting, but their two-stage pipeline introduces redundant computation and heavily relies on SAM's capabilities, limiting their applicability to 3D medical images. Neuralizer~\cite{czolbe2023neuralizer} develop general tools on diverse neuroimaging tasks, like super-resolution denosing, etc. Recent works introduced specialized architectures for in-context segmentation~\cite{meng2024segic}. For example, UniverSeg~\cite{butoi2023universeg} is designed for in-context medical image segmentation, and Tyche~\cite{rakic2024tyche} incorporated a stochastic inference. While these methods demonstrate promising capability on novel classes, they face two critical limitations. First, they show suboptimal performance compared to task-specific models on the training distribution. Second, they suffer from computational inefficiencies as they can only segment one anatomical class per forward pass, requiring multiple passes for multi-class segmentation. Meanwhile they must re-encode context examples for each query image even when using the same reference examples repeatedly. This becomes particularly problematic in high-throughput scenarios where multiple query images need to be processed. In contrast, Iris shows better performance and efficiency. An appealing design of our context task encoding module is to decouple task definition from inference, enabling the encoding of task from reference pairs into task tokens that can be efficiently reused across any number of query images, meanwhile multi-class segmentation can be done within a single forward pass.

The selection of appropriate context examples impacts the performance of in-context learning. Current methods~\cite{zhang2023makes} employ image-level retrieval strategies using global image embeddings to find better references. However, this approach faces significant challenges in medical image analysis where each image contains multiple classes including structures like organs, tissues, and lesions. Image-level retrieval inevitably averages features across all structures, leading to a suboptimal reference selection. To address this limitation, Iris introduces an object-level context selection mechanism that enables fine-grained matching of individual classes, focusing on more precise and class-specific reference selection compared to image-level approaches~\cite{zhang2023makes}.

%% file: sec/3_method.tex
\section{Method}
\label{sec:method}

\subsection{Problem Definition}
Traditional segmentation approaches follow a task-specific paradigm, where each model $f_{\theta_t}$ is trained for a specific segmentation task $t$. Given a dataset $\mathcal{D}_t = \{(\boldsymbol{x}_{t}^{i}, \boldsymbol{y}_{t}^{i})\}_{i=1}^{N_t}$ containing $N_t$ image-label pairs, the model learns a direct mapping $f_{\theta_t}: \mathcal{X} \rightarrow \mathcal{Y}$ from the image space $\mathcal{X}$ to the segmentation mask space $\mathcal{Y}$, such that for an image $\boldsymbol{x}_t$, the predicted segmentation mask is given by $\boldsymbol{y}_t = f_{\theta_t}(\boldsymbol{x}_t)$.

In contrast, we formulate a \textit{in-context medical image segmentation} framework. Given a support set $\mathcal{S} = \{(\boldsymbol{x}_s^i, \boldsymbol{y}_s^i)\}_{i=1}^n \in (\mathcal{X} \times \mathcal{Y})^n$ containing $n$ reference image-label pairs and a query image $\boldsymbol{x}_q \in \mathcal{X}$, a single model $f_\theta$ predicts the segmentation mask $\boldsymbol{y}_q$ for the query image conditioned on $\mathcal{S}$:
\begin{equation}
    \boldsymbol{\hat{y}}_q = f_\theta(\boldsymbol{x}_q; \mathcal{S}) = f_\theta(\boldsymbol{x}_q; \{(\boldsymbol{x}_s^i, \boldsymbol{y}_s^i)\}_{i=1}^n)
\end{equation}
For multi-class segmentation tasks, we decompose the problem into multiple binary segmentation tasks.

\subsection{Iris Architecture}
In Figure~\ref{fig:framework}, Iris introduces a novel in-context learning architecture that decouples task encoding from segmentation inference. This design comprises two key components: (1) a task encoding module that distills task-specific information from reference examples into compact task embeddings, and (2) a mask decoding module that leverages these task embeddings to guide query image segmentation.

\subsubsection{Task Encoding Module}
Given a reference 3D image-label pair $(\boldsymbol{x}_s, \boldsymbol{y}_s) \in \mathbb{R}^{D \times H \times W} \times \{0,1\}^{D \times H \times W}$, our task encoding module extracts task representations through two parallel streams to extract comprehensive task representations.

\noindent\textbf{Foreground feature encoding.} Medical data volumes present unique challenges in feature extraction due to the presence of fine boundary details and anatomical structures spanning only a tiny portion of voxels. Direct feature pooling at downsampled resolution can lead to information loss or complete disappearance of these critical regions of interest (ROIs). To address this hurdle, we opt in a high-resolution foreground feature encoding process. Given features $\boldsymbol{F}_s \in \mathbb{R}^{C \times d \times h \times w}$ extracted by the encoder $E$, where $d=D/r, h=H/r, w=W/r$ are downsampled dimensions with ratio $r$, we compute the foreground embedding by:
\begin{equation}
\boldsymbol{T}_f = \text{Pool}(\text{Upsample}(\boldsymbol{F}_s) \odot \boldsymbol{y}_s) \in \mathbb{R}^{1 \times C}
\end{equation}
where $\text{Upsample}(\boldsymbol{F}_s) \in \mathbb{R}^{C \times D \times H \times W}$ restores features to the original resolution. By applying the original high-resolution mask $\boldsymbol{y}_s$ directly to the upsampled features, we ensure a precise capture of fine anatomical details and small structures that are vital for medical object segmentation.

\noindent\textbf{Contextual feature encoding.} The above encoding process extracted foreground features, but lacks important global context information. We encode these contextual information using learnable query tokens. To efficiently process high-resolution features while managing memory constraints,  we employ strategy similar to sub-pixel convolution~\cite{shi2016real}. For feature map $\boldsymbol{F}_s$, we first expand spatial dimensions while reducing channels:
\begin{equation}
\boldsymbol{F}'_s = \text{PixelShuffle}(\boldsymbol{F}_s) \in \mathbb{R}^{C/r^3 \times D \times H \times W}
\end{equation}
After concatenating with the binary mask $\boldsymbol{y}_s$, we apply a $1\times1\times1$ convolution and PixelUnshuffle to return to the original feature resolution:
\begin{equation}
\hat{\boldsymbol{F}}_s = \text{PixelUnshuffle}(\text{Conv}(\text{Concat}[\boldsymbol{F}'_s, \boldsymbol{y}_s])) \in \mathbb{R}^{C \times d \times h \times w}
\end{equation}
This approach permits a memory-efficient, high-resolution, feature-mask fusion. The merged features $\hat{\boldsymbol{F}}_s$ then interact with $m$ learnable query tokens through cross-attention and self-attention layers to produce contextual embedding $\boldsymbol{T}_c \in \mathbb{R}^{m \times C}$. The final task embedding combines both aspects: $\boldsymbol{T} = [\boldsymbol{T}_f; \boldsymbol{T}_c] \in \mathbb{R}^{(m+1) \times C}$.

For multi-class segmentation, we generate separate task embeddings for each category in $\boldsymbol{y}_s$. This setting maintains a strong efficiency as the computationally intensive feature extraction is shared across classes while the task encoding module remains lightweight.

\subsubsection{Mask Decoding Module}
The decoder $D$ employs a query-based architecture~\cite{cheng2022masked} that efficiently handles both single and multi-class segmentation tasks. For a query image with features $\boldsymbol{F}_q \in \mathbb{R}^{C \times d \times h \times w}$, the task encoding module generates class-specific embeddings $\boldsymbol{T}^k \in \mathbb{R}^{(m+1) \times C}$ for each class $k$ defined in reference image-label pairs. These embeddings are concatenated into a combined task representation $\boldsymbol{T} = [\boldsymbol{T}^1; \boldsymbol{T}^2; ...; \boldsymbol{T}^K] \in \mathbb{R}^{K(m+1) \times C}$, where $K$ is the number of target classes and $K=1$ for single-class segmentation. The bidirectional cross-attention mechanism processes this representation:
\begin{equation}
\boldsymbol{F}'_q, \boldsymbol{T}' = \text{CrossAttn}(\boldsymbol{F}_q, \boldsymbol{T})
\end{equation}
where $\boldsymbol{F}'_q$ and $\boldsymbol{T}'$ are the updated features. This mechanism enables effective information exchange between class-specific task guidance and query image features. The final segmentation mask is predicted in a single forward pass:
\begin{equation}
\boldsymbol{\hat{y}}_q = D(\boldsymbol{F}'_q, \boldsymbol{T}') \in \{0,1\}^{K \times D \times H \times W}
\end{equation}

\subsubsection{Training} We train Iris in an end-to-end manner using episodic training to simulate in-context learning scenarios (details in supplementary). Each training episode consists of sampling reference-query pairs from the same dataset, computing task embeddings from the reference pair, and final predicting segmentation for the query image. The model is optimized using a combination of Dice and cross-entropy losses: $\mathcal{L}_{seg} = \mathcal{L}_{dice} + \mathcal{L}_{ce}$. To enhance generalization, we employ data augmentation on both query and reference images, add random perturbation to query images to simulate imperfect references, and randomly drop classes in multi-class datasets to encourage independent class-wise task encoding.

\begin{figure}[t]
\begin{center}
\includegraphics[width=\columnwidth]{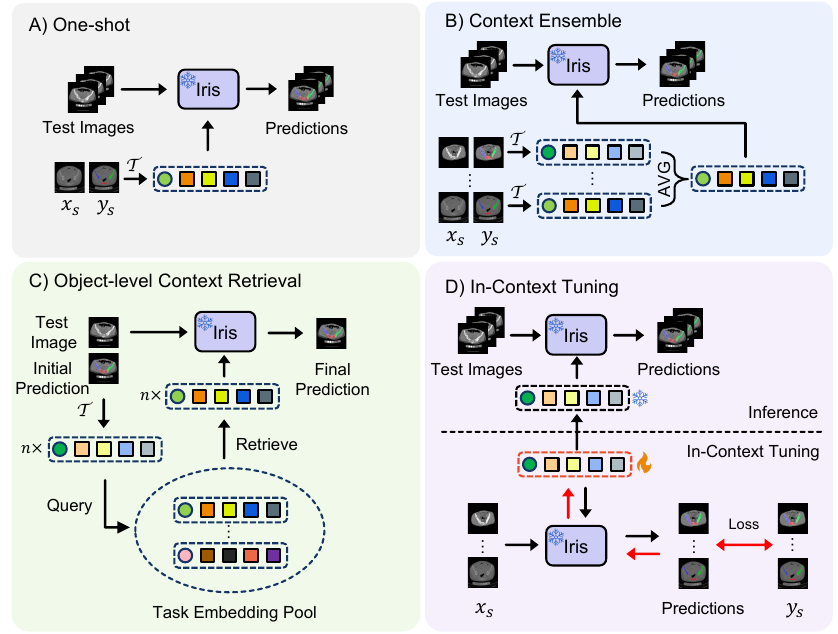}
\end{center}
\vspace{-1em}
\caption{Iris's flexible inference strategies. The red arrows indicates gradient backpropagation.}
\label{fig:inference_strategy}
\vspace{-1em}
\end{figure}

\begin{table*}[t]
\centering
\caption{Comparison of segmentation performance across different in-distribution datasets. Values represent mean Dice scores (\%).}
\label{tab:id_results}
\resizebox{\textwidth}{!}{
\begin{tabular}{l|cccccccccccc|c}
\toprule
\multirow{2}{*}{Method} & \multicolumn{12}{c|}{Dataset} & \multirow{2}{*}{AVG} \\
\cline{2-13}
& AMOS & AMOS & Auto & BCV & Brain & CHAOS & KiTS & LiTS & MnM & StructSeg & StructSeg & CSI-Wat & \\
& CT & MR & PET & & & & Tumor &Tumor & & H\&N & Tho & & \\
\midrule
\multicolumn{14}{l}{\textit{Task-specific Model (Upper Bound)}} \\
nnUNet & 88.67 & 85.42 & 67.21 & 83.38 & 94.12 & 91.13 & 81.72 & 63.11 & 85.59 & 78.17 & 88.53 & 91.11 & 83.18 \\
\midrule
\multicolumn{14}{l}{\textit{Multi-task Universal Model (Upper Bound)}} \\
Clip-driven & 88.95 & 86.41 & 70.01 & 85.03 & 95.06 & 91.71 & 82.73 & 65.43 & 86.12 & 78.44 & 89.27 & 90.98 & 84.18 \\
UniSeg & 89.11 & 86.58 & 70.09 & 85.42 & 95.29 & 91.83 & 82.99 & 65.87 & 86.29 & 78.72 & 89.42 & 91.23 & 84.40 \\
Multi-Talent & 89.15 & 86.58 & 70.89 & 85.20 & 95.77 & 91.38 & 82.32 & 65.53 & 86.30 & 80.09 & 89.09 & 91.32 & 84.47 \\
\midrule
\multicolumn{14}{l}{\textit{Positional Prompt}} \\
SAM & 22.23 & 17.82 & 20.10 & 23.34 & 20.51 & 20.01 & 18.21 & 12.08 & 10.23 & 17.23 & 24.81 & 13.20 & 17.97 \\
SAM-Med 2D & 50.12 & 48.66 & 38.03 & 50.32 & 35.28 & 50.32 & 30.23 & 23.27 & 40.33 & 39.32 & 63.87 & 34.87 & 40.58 \\
SAM-Med 3D & 79.19 & 76.18 & 67.14 & 79.89 & 42.29 & 84.79 & 79.32 & 32.93 & 52.67 & 68.83 & 83.56 & 74.23 & 68.42 \\
\midrule
\multicolumn{14}{l}{\textit{In-Context}} \\
SegGPT & 45.37 & 51.78 & 48.29 & 49.78 & 85.27 & 63.72 & 40.78 & 35.98 & 74.12 & 40.28 & 67.28 & 85.59 & 57.35 \\
UniverSeg & 57.24 & 52.43 & 47.23 & 45.26 & 87.76 & 60.46 & 45.72 & 36.21 & 75.24 & 42.98 & 66.95 & 86.68 & 58.68 \\
Tyche-IS & 59.57 & 54.78 & 50.98 & 47.67 & 89.28 & 62.73 & 49.27 & 37.02 & 78.92 & 45.33 & 69.89 & 88.99 & 61.20 \\
\rowcolor{lightgray} Iris (ours) & 89.56 & 86.70 & 70.02 & 85.73 & 96.04 & 91.85 & 81.54 & 65.02 & 86.08 & 80.36 & 89.42 & 91.97 & 84.52 \\
\bottomrule
\end{tabular}
}
\vspace{-1em}
\end{table*}

\subsection{Flexible Inference Strategies}
After training, Iris supports multiple inference strategies suitable for different practical scenarios (see in Figure~\ref{fig:inference_strategy}).

\noindent\textbf{Efficient one-shot inference.} With just one reference example, Iris first encodes the task into compact embeddings $\boldsymbol{T}$ that can be stored and reused across multiple query images. Unlike major in-context learning methods to recompute contextual information for each query image, our design greatly eliminates redundant computation. Moreover, Iris can segment multiple classes in a single forward pass, contrasting with methods (e.g., UniverSeg~\cite{butoi2023universeg}) that require separate passes per class. The minimal storage requirement of these embeddings makes Iris particularly desirable for large-scale data processing pipelines.

\noindent\textbf{Context ensemble.}
For tasks with multiple reference examples, Iris supports context ensemble for improving performance. We compute task embeddings for each example and average them to create a more robust task representation. This simple averaging strategy combines information from multiple references while maintaining computational efficiency. We extent context ensemble for classes seen during training. Specifically, we maintain a class-specific memory bank that continuously updates task embeddings through exponential moving average (EMA) during the training process. This memory bank stores representative task embeddings for each seen class, enabling direct segmentation for seen classes during inference without requiring context encoding.

\noindent\textbf{Object-level context retrieval.} For multi-class segmentation with a pool of reference examples, conventional approaches typically employ image-level retrieval using global embeddings to select semantically similar references~\cite{zhang2023makes}. However, this strategy is suboptimal for medical images where multiple anatomical structures coexist, as global embeddings average features across all structures. To enable more precise reference selection, we propose an object-level (class-level) context retrieval strategy. Our approach first encodes class-specific task embeddings for each reference example through our task encoding module - for a reference image with $n$ anatomical classes, we encode $n$ separate task embeddings. For a query image, we obtain initial object segmentation masks using task embeddings from a randomly selected reference. These initial masks are then used to encode $n$ class-specific query task embeddings, which are compared with corresponding reference embeddings in the pool using cosine similarity to select the most similar reference for each class independently. This fine-grained matching allows different structures within the same query image to find their most appropriate references, leading to more accurate segmentation compared to image-level approaches.

\noindent\textbf{In-context tuning.} For scenarios requiring adaptation without a full model fine-tuning, Iris offers a lightweight tuning strategy by optimizing only the task embeddings while keeping the model parameters fixed. This tuning process minimizes the segmentation loss between model predictions and ground truth by updating the task embeddings through the gradient descent. In particular, the optimized embeddings can then be stored and reused for similar cases, offering a practical balance between adaptation capability and computational efficiency.

%% file: sec/4_experiment.tex
\section{Experiment}

\begin{table*}[t]
\centering
\caption{Out-of-distribution comparison on held-out datasets, including generalization capability and performance on unseen classes. Values represent mean Dice scores (\%). All in-context models use one-shot inference.}
\label{tab:ood_results}
\setlength{\tabcolsep}{1.4em}
\resizebox{0.9\textwidth}{!}{ 
\begin{tabular}{l|ccccc|cc}
\toprule
\multirow{2}{*}{Method} & \multicolumn{5}{c|}{Generalization} & \multicolumn{2}{c}{Unseen Classes} \\
\cline{2-8}
& ACDC & SegTHOR & CSI-inn & CSI-opp & CSI-fat & MSD Pancreas & Pelvic \\
\midrule
\multicolumn{8}{l}{\textit{Supervised Upper Bound}} \\
nnUNet & 90.97 & 89.78 & 91.23 & 91.04 & 90.13 & 54.56 & 94.73 \\
\midrule
\multicolumn{8}{l}{\textit{Task-specific Model}} \\
nnUNet-generalize & 82.06 & 76.92 & 55.24 & 85.19 & 0.23 & -- & -- \\
\midrule
\multicolumn{8}{l}{\textit{Multi-task Universal Model}} \\
CLIP-driven & 84.72 & 78.23 & 59.73 & 86.73 & 1.47 & -- & -- \\
UniSeg & 84.98 & 78.56 & 60.02 & 86.13 & 1.52 & -- & -- \\
Multi-Talent & 83.79 & 78.45 & 58.29 & 87.01 & 1.95 & -- & -- \\
\midrule
\multicolumn{8}{l}{\textit{Positional Prompt}} \\
SAM-Med2D & 42.23 & 52.37 & 29.23 & 32.71 & 10.91 & 10.37 & 35.71 \\
SAM-Med3D & 51.49 & 68.97 & 45.32 & 68.72 & 23.93 & 15.83 & 53.61 \\
\midrule
\multicolumn{8}{l}{\textit{In-context}} \\
SegGPT & 73.82 & 60.98 & 59.87 & 77.62 & 35.27 & 10.67 & 55.92 \\
UniverSeg & 72.43 & 54.75 & 63.48 & 85.32 & 52.48 & 10.28 & 57.81 \\
Tyche-IS & 74.91 & 56.75 & 64.23 & 87.13 & \textbf{55.75} & 11.97 & 61.92 \\
\rowcolor{lightgray} Iris (ours) & \textbf{86.45} & \textbf{82.77} & \textbf{64.44} & \textbf{89.13} & 47.78 & \textbf{28.28} & \textbf{69.03} \\
\bottomrule
\end{tabular}
}
\vspace{-1em}
\end{table*}

\subsection{Experimental Setup}
We evaluate Iris across three key dimensions: in-distribution performance on trained tasks, out-of-distribution generalization to different domains, and adaptability to novel anatomical classes. Additional experiments analyze Iris's computational efficiency, inference strategies, and architectural design choices.

\noindent\textbf{Dataset.}
Our training data comprises 12 public datasets~\cite{bcv,bilic2019liver,heller2019kits19,ji2022amos,structseg,CHAOS2021,campello2021multi,rodrigue2012beta,gatidis2022whole,martin2023deep} spanning diverse body regions (head, chest, abdomen), modalities (CT, MRI, PET), and clinical targets (organs, tissues, lesions), split into 75\%/5\%/20\% for train/validation/test. For out-of-distribution evaluation, we use 5 held-out datasets: ACDC~\cite{bernard2018deep}, SegTHOR~\cite{lambert2020segthor}, and three MRI modalities from IVDM3Seg~\cite{ivdm3seg} to evaluate robustness against domain shift; MSD Pancreas (Tumor)~\cite{antonelli2022medical} and Pelvic1K (Bone)~\cite{liu2021deep} datasets are used for novel class adaptation. We randomly select 20\% samples from held-out sets for testing. Detailed dataset information is provided in supplementary materials.

\noindent\textbf{Baseline Models.} 
We compare against four categories of methods:
(1) Task-specific models: nnUNet~\cite{isensee2021nnu};
(2) Universal models: CLIP-driven model~\cite{liu2023clip}, UniSeg~\cite{ye2023uniseg} and Multi-Talent~\cite{ulrich2023multitalent};
(3) Foundation models: SAM~\cite{kirillov2023segment} and its medical variants, SAM-Med2D~\cite{cheng2023sam}, SAM-Med3D~\cite{wang2024sam};
(4) In-context learning methods: SegGPT~\cite{wang2023seggpt}, UniverSeg~\cite{butoi2023universeg}, and Tyche-IS~\cite{rakic2024tyche}.
All models are trained on our curated dataset, except SAM, with 2D models trained on extracted slices and 3D models with 3D volumes. For SAM-based methods, we simulate user interactions using ground-truth labels during training and evaluation.

\noindent\textbf{Implementation Details.}
Iris uses a 3D UNet encoder trained from scratch with one-shot learning strategy. We employ the Lamb optimizer~\cite{you2019large} with exponential learning rate decay (base lr=$2\times10^{-3}$, weight decay=$1\times10^{-5}$), training for 80K iterations with batch size 32 and 2K warm-up iterations. Data augmentation includes random cropping, affine transformations, and intensity adjustments. Training and inference use $128\times128\times128$ volume size.

\subsection{Comparison with the state-of-the-art}

\noindent\textbf{Results on in-distribution classes.}
We evaluate Iris's performance on twelve diverse medical datasets used during training. As shown in Table \ref{tab:id_results}, Iris achieves state-of-the-art performance with an average Dice score of 84.52\%, matching or exceeding task-specific and multi-task models that are optimized for fixed tasks.
Existing adaptive approaches show significant limitations. SAM-based methods perform poorly due to their strong reliance on simple positional prompts, with the large performance gap between their 2D and 3D variants (40.58\% vs. 68.42\%) highlighting the importance of 3D context. Existing in-context learning methods, like SegGPT, UniverSeg, (best: 61.20\%) struggle particularly with 3D tasks like AMOS and LiTS due to their 2D architecture, though performing better on 2D-friendly tasks like MnM and CSI-Wat. In contrast, Iris's 3D architecture and efficient task encoding enables consistent high-level performance across all tasks while maintaining its adaptability to unseen novel anatomical classes.

\noindent\textbf{Results on OOD generalization.} We evaluate out-of-distribution (OOD) performance on five held-out datasets spanning two types of distribution shifts: cross-center variation (ACDC, SegTHOR) and cross-modality adaptation (CSI variants). In Table \ref{tab:ood_results}, Iris demonstrates superior performance across all scenarios, particularly excelling in challenging 3D tasks and large domain shifts.

Both task-specific and multi-task universal models show performance degradation, especially failing catastrophically on CSI-fat with a significant domain gap. While SAM-based methods demonstrate their resilience to domain shifts through strong prior knowledge injected from user interactions, their performance remains limited on the volumetric data.
In-context learning methods retain a good performance with cross-modality adaptation (e.g. on CSI-fat), benefiting from the domain-specific knowledge provided by reference examples. However, a 2D-slice-based architecture (e.g., UniverSeg and Tyche) limits its capability on 3D tasks like SegTHOR. In contrast, Iris's task encoding module efficiently extracts and utilizes 3D domain-specific information from the reference examples.

\noindent\textbf{Results on novel classes.} To measure the adaptation performance to completely unseen anatomical structures, we evaluate on MSD Pancreas Tumor and Pelvic datasets. Using only one reference example, Iris achieves 28.28\% on MSD Pancreas Tumor and 69.03\% on Pelvic segmentation, substantially outperforming other adaptive methods (the best competitor: 11.97\% and 61.92\% respectively). This performance gain is particularly notable given that traditional task-specific models and multi-task models can not handle these novel classes without retraining. These findings demonstrate Iris's strong capability in learning from very limited examples while maintaining meaningful segmentation quality on previously unseen anatomical structures.

\noindent\textbf{Efficiency comparison}
 We analyze computational efficiency for segmenting $m$ query images with $n$ classes using $k$ reference pairs. Iris achieves superior efficiency through two key designs: decoupling task extraction from inference and handling multiple classes in a single forward pass. This results in complexity of $O(k + m)$ compared to $O(kmn)$ in methods like UniverSeg that process each class separately and recompute reference features for every query.

Table~\ref{tab:complexity} compares real-world inference time and memory usage across methods. While UniverSeg's slice-by-slice processing leads to significant overhead with multiple reference slices, and SAM-Med3D requires iterative user interactions (interaction-time not included), Iris efficiently processes entire 3D volumes all at once. For a scenario of segmenting 10 query volumes with 15 classes using one reference volume, Iris completes in 2 seconds. This efficiency advantage grows with more context examples due to Iris's decoupled architecture eliminating redundant reference processing.

\begin{table}[t]
\centering
\caption{Comparison of computational complexity. Empirical measurements of computation on one NVIDIA A100 GPU. We inference with 10 query images and one reference image from AMOS CT dataset with 15 classes. The image size is processed to $128\times128\times128$ for inference.}
\label{tab:complexity}
\resizebox{0.9\columnwidth}{!}{
\begin{tabular}{lccc}
\toprule
Method & Inference Time (s) & Memory (GB) & Parameters (M) \\
\midrule
UniverSeg-1 & 659.4 & 2.1 & 1.2 \\
UniverSeg-128 & 1030.2 & 12.1 & 1.2 \\
SAM-Med2D & 648.4 & 1.8 & 91.1 \\
SAM-Med3D & 15.2 & 2.9& 100.5 \\
\rowcolor{lightgray}Iris (Ours) & 2.0 & 7.4 & 69.4 \\
\bottomrule
\end{tabular}
}
\vspace{-1em}
\end{table}

\subsection{Analysis}

\begin{figure}[t]
\begin{center}
\includegraphics[width=0.7\columnwidth]{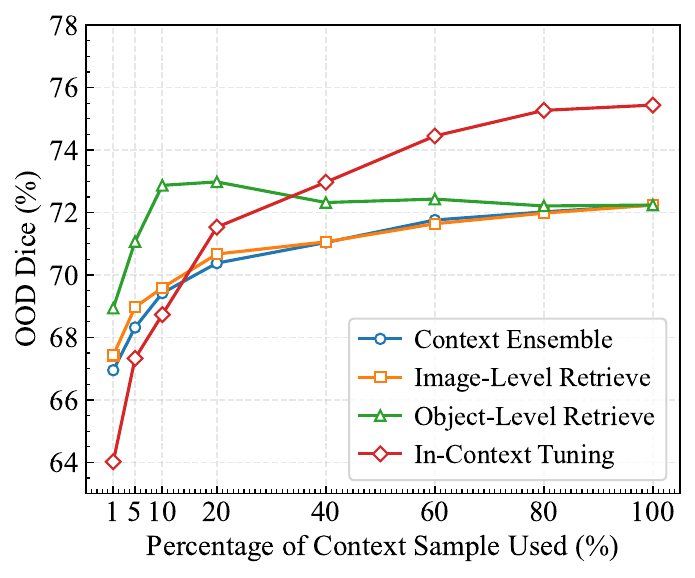}
\end{center}
\vspace{-1em}
\caption{Analysis of different inference strategies.}
\label{fig:inference_comparison}
\vspace{-1em}
\end{figure}

\noindent\textbf{Inference strategy.} Figure~\ref{fig:inference_comparison} compares our four inference strategies. In this experiment, we maintain a pool of all available context examples and evaluate each strategy's performance as follows.

Context ensemble randomly selects and averages task embeddings from a percentage of the context pool. When using only one context example (1\%), it operates as one-shot inference. Performance of context ensemble keeps improving with more context examples and eventually saturates. This strategy is appealing as task embeddings can be precomputed and ensembled into a single robust embedding, enabling inference speed comparable to regular segmentation models.

Both image and object-level retrieval strategies access the entire context pool but utilize only the top-$k$ percent most similar examples as references. While image-level retrieval~\cite{zhang2023makes} compares whole-image features and uses all task embeddings from the same retrieved images, object-level retrieval enables more precise reference selection by matching individual classes. Notably, object-level retrieval surpasses full context ensemble performance when using fewer references (e.g., top 10-20\%), as it selectively chooses the most relevant examples for each class rather than averaging all available contexts. To validate robustness to initial context selection, we conducted experiments for 10 times with random selection using different percentages of context samples (1\%, 5\%, 10\%), achieving consistent performance (mean and standard deviation: $68.94\pm 0.83$, $71.07\pm 0.27$, $72.87\pm 0.10$ respectively). This strategy is particularly valuable in clinical settings with large patient databases, where retrieving similar cases as references can enhance segmentation accuracy.

In-context tuning optimizes task embeddings initialized from a random reference. While showing a lower performance with limited samples due to overfitting, it achieves positive results with sufficient tuning data. This approach is well suited for scenarios with both a large context pool and available computational resources for fine-tuning.

Overall, Iris offers usable strategies pertinent to different real-world scenarios. Object-level retrieval is designed for high accuracy while requiring access to a large context pool, e.g. a database of previous patient records. Context ensemble offers a strong efficiency of response time. Finally, in-context tuning is applicable when computational resources and sufficient data support are available.

\begin{figure}[t]
\begin{center}
\includegraphics[width=0.9\columnwidth]{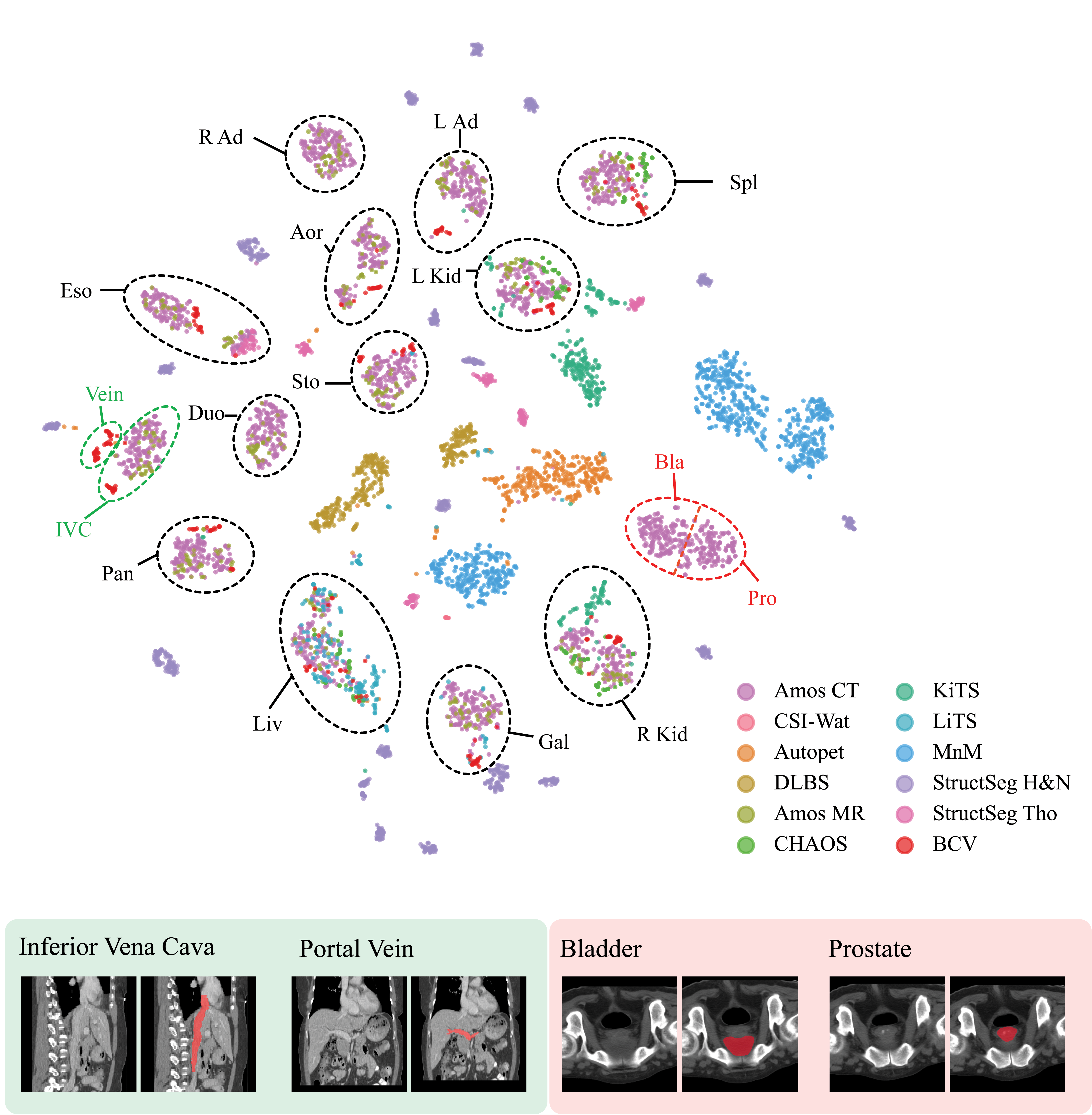}
\end{center}
\vspace{-1em}
\caption{\textbf{Top}: Visualizing the task embedding with t-SNE. The color represents dataset, the circle and marks are the classes of the embeddings. \textbf{Bottom}: Examples of the similar tasks revealed by the t-SNE result. }
\label{fig:task_embedding_vis}
\vspace{-1em}
\end{figure}

\noindent\textbf{Task embedding analysis.} Iris's task encoding module discovers meaningful anatomical relationships without explicit anatomical supervision, learning solely from binary segmentation masks. From Figure~\ref{fig:task_embedding_vis}, the t-SNE visualization of task embeddings reveals natural clustering of anatomical structures that transcends dataset boundaries and imaging modalities. For example, abdominal organs cluster together despite originating from different datasets and modalities (e.g., AMOS-CT, BCV in CT; AMOS-MR, CHAOS in MRI).

We find that feature embeddings capture clinically meaningful anatomical similarities that were never explicitly taught (Figure~\ref{fig:task_embedding_vis}, bottom). Blood vessels like the Inferior Vena Cava (IVC) and Portal/Splenic veins cluster nearby, reflecting their shared tubular structure and similar contrast enhancement patterns in CT. Similarly, the bladder and prostate embeddings show proximity due to their shared soft-tissue characteristics and adjacent anatomical locations. This emergent organization of anatomical concepts demonstrates Iris's ability to automatically distill fundamental anatomical relationships across different segmentation tasks, making it particularly robust for adapting to new anatomical structures.

\begin{figure}[t]
\begin{center}
\includegraphics[width=\columnwidth]{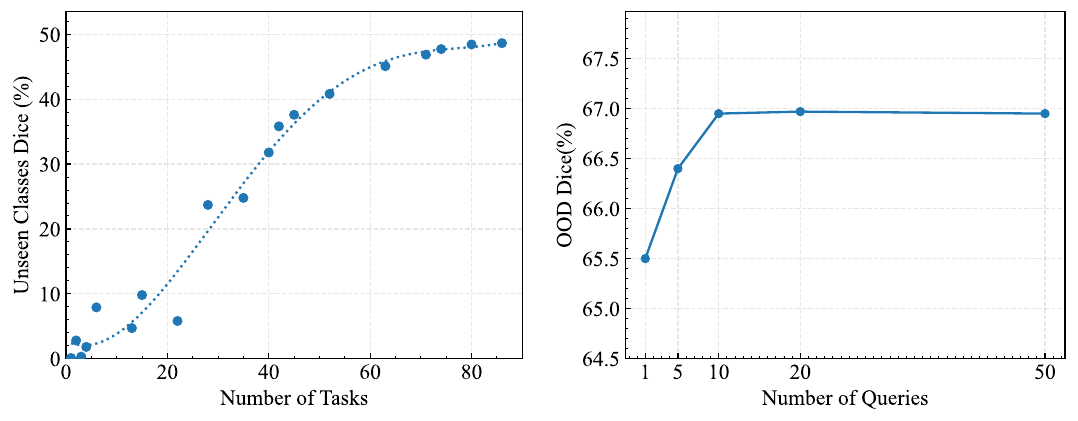}
\end{center}
\vspace{-1em}
\caption{\textbf{Left}: Number of tasks used for training v.s. Performance on unseen classes. \textbf{Right}: Ablation on the number of queries. }
\label{fig:ablation}
\vspace{-1em}
\end{figure}

\noindent\textbf{Generalization performance vs task quantity.}
We investigate how training data diversity affects Iris's generalization by varying the number of training tasks. From Figure~\ref{fig:ablation} (left), the performance on held-out datasets consistently improves with more training tasks, particularly when the training subset encompasses diverse anatomical structures and imaging modalities. We recognize that models trained on datasets spanning body regions (e.g., brain, chest, and abdomen) show a stronger generalization compared to those trained on narrower anatomical ranges. This finding suggests that an exposure to diverse anatomical patterns is necessary towards more robust and transferable feature learning.

\noindent\textbf{Ablation study.}
Table~\ref{tab:ablation} analyzes three key components of Iris. High-resolution processing proves crucial for small structures, dramatically improving their segmentation performance from 62.13\% to 78.92\%. While each component contributes independently, their combination achieves the best results across all metrics, demonstrating substantial improvements over partial implementations. In Figure~\ref{fig:ablation} (right), performance improves with more query tokens but saturates at 10 tokens, which we adopt in our final model to balance performance and efficiency.

\begin{table}[t]
\centering
\caption{Ablation study of different components in Iris. High-Res: high-resolution feature processing; Foreground: foreground feature pooling; Query: query-based contextual encoding.}
\label{tab:ablation}
\resizebox{\columnwidth}{!}{
\begin{tabular}{cccccc}
\toprule
High-Res & Foreground & Query & In-dist (Avg) & In-dist (Small) & Out-of-dist \\
\midrule
         & \checkmark & \checkmark & 82.10 & 62.13 & 62.00 \\
\checkmark & \checkmark   & & 82.47 & 78.92 & 65.93 \\
\checkmark &  &\checkmark           & 82.06 & 77.53 & 64.13 \\
\rowcolor{lightgray}
\checkmark & \checkmark & \checkmark & \textbf{84.52} & \textbf{80.36} & \textbf{66.95} \\
\bottomrule
\end{tabular}
}
\vspace{-1em}
\end{table}

%% file: sec/5_discussion.tex
\section{Discussion and Conclusion}

We introduce Iris as a novel in-context learning framework that enables versatile 3D medical image segmentation through only reference examples. Given just one image-label pair as a reference, Iris can segment arbitrary target classes in test images without any model modification or retraining. Iris reveals strong performance on in-distribution tasks across 12 diverse datasets. Iris's performance is particularly evident to distribution shifts and novel unseen classes on 7 held-out test datasets. The key design of Iris is a decoupled architecture that enables efficient 3D medical image processing and single-pass multi-class segmentation. Iris's inference strategies are suitable for different practical scenarios, from efficient context ensemble-based data processing, high-accuracy object-level context retrieval, to in-context finetuning. Further, Iris's task encoding module offers an appealing means to automatically discover meaningful anatomical relationships purely from segmentation masks, allowing knowledge transfer across different tasks and imaging modalities without explicit anatomical supervision.

\noindent\textbf{Limitations and future work.} While Iris demonstrates promising capabilities, several challenges remain to explore. The diversity of training tasks could impact the out-of-distribution generalization, suggesting a critical need for automated methods to create diverse tasks without manual annotation. Although Iris shows strong adaptability to novel tasks, there remains a performance gap with supervised upper bounds in certain scenarios. Future investigation will focus on narrowing this gap and expanding both training and evaluation schemes to cover a broader spectrum of medical imaging applications.

%% file: sec/X_suppl.tex
\clearpage
\setcounter{page}{1}
\maketitlesupplementary

\section{Dataset Details}

This section provides comprehensive information about our experimental datasets, including data characteristics, annotation details, acquisition protocols, and their roles in our experimental setup. We describe both the datasets used for upstream training and those held-out for out-of-distribution evaluation.

\noindent\textbf{Multi-organ Abdominal Collection (AMOS).} AMOS~\cite{ji2022amos} represents a comprehensive multi-modal dataset from Longgang District People's Hospital, featuring 500 CT and 100 MRI scans from 600 patients with abdominal abnormalities. Acquired across eight different scanner platforms, the dataset provides annotations for 15 anatomical structures, including major abdominal organs, vessels, and reproductive organs: spleen, right kidney, left kidney, gallbladder, esophagus, liver, stomach, aorta, inferior vena cava, pancreas, right adrenal gland, left adrenal gland, duodenum, bladder, and prostate/uterus. The CT portion offers 200 training and 100 validation scans, while the MRI section provides 40 training and 20 validation scans. We employ both modalities in upstream training, using a 95/5 split for training/validation using the official training set, while using the official validation set for evaluation. Note that the MRI validation set lacks bladder and prostate annotations, limiting MRI segmentation to 13 structures.

\noindent\textbf{Whole-body PET/CT Collection (AutoPET).} AutoPET~\cite{gatidis2022whole} represents a comprehensive collection of 1014 whole-body FDG-PET/CT studies, balanced between 501 cases with confirmed malignancies (lymphoma, melanoma, NSCLC) and 513 negative control cases. All scans include both PET and CT modalities, making it valuable for multi-modal analysis. We maintain patient-level data integrity with a 75\%/5\%/20\% split for training, validation, and testing.

\noindent\textbf{Abdominal CT from Multi-Atlas (BCV).} The BCV~\cite{bcv}  collection consists of 50 abdominal CT scans obtained during routine clinical care at Vanderbilt University Medical Center (VUMC). Of these, 30 scans are publicly accessible with volumetric annotations of 13 abdominal organs created using MIPAV software. The annotated structures encompass major organs and vessels including the liver, kidneys (left/right), pancreas, spleen, gallbladder, esophagus, stomach, aorta, inferior vena cava, portal and splenic veins, and adrenal glands (left/right). Notable is the occasional absence of right kidney or gallbladder annotations in some patients. For our upstream training pipeline, we implement a 75\%/5\%/20\% split of the available data for training, validation, and testing respectively.

\noindent\textbf{Brain Aging Study Collection (Brain)~\cite{rodrigue2012beta}.} Part of the Dallas Lifespan Brain Study, this dataset aims to understand cognitive function changes across adult life, particularly focusing on early indicators of Alzheimer's Disease progression. Our analysis utilizes 213 T1-weighted MRI scans, annotated for three key brain tissue types: cerebrospinal fluid, gray matter, and white matter. Following established protocols~\cite{rao2022improving}, we distribute the scans into 129 training, 43 validation, and 43 testing cases.

\noindent\textbf{Abdominal MRI Collection (CHAOS).} CHAOS~\cite{CHAOS2021} focuses on precise abdominal organ segmentation in magnetic resonance imaging. The dataset features multi-sequence MRI scans (T1-in-phase, T1-out-phase, T2-SPIR) from 20 patients, with annotations of four major abdominal organs: liver,  left kidney, right kidney, and spleen. Each MR sequence is treated as an independent image for analysis purposes, while maintaining patient-level data splits of 75/5/20 for training, validation, and testing to prevent data leakage.

\noindent\textbf{Kidney Tumor Dataset (KiTS19)~\cite{heller2019kits19}.} Sourced from the University of Minnesota Medical Center between 2010-2018, KiTS19 comprises CT scans and treatment outcomes from 300 kidney tumor patients who underwent nephrectomy procedures. The publicly available portion includes 210 cases, while 90 remain private for evaluation purposes. We incorporate this dataset into our upstream training using a 75\%/5\%/20\% of the 210 training cases for training/validation/testing.

\noindent\textbf{Liver Cancer Imaging Collection (LiTS)~\cite{bilic2019liver}.} This dataset encompasses 201 abdominal CT scans (131 training, 70 testing) gathered from seven prominent medical institutions including centers in Munich, Nijmegen, Montreal, Tel Aviv, and Strasbourg. The collection features patients with various liver malignancies, including primary hepatocellular carcinoma and metastases from colorectal, breast, and lung cancers. The scans exhibit diverse tumor characteristics and contrast enhancement patterns, captured both pre- and post-treatment using various CT protocols. Annotations include detailed tumor delineation alongside broader liver segmentation. We utilize the 131 public training cases with a 75\%/5\%/20\% split for our upstream training framework.

\noindent\textbf{Cardiac MRI Dataset (M\&Ms).} The M\&Ms~\cite{campello2021multi} dataset represents a diverse cardiac imaging collection from the MICCAI 2020 Challenge, featuring scans from patients with cardiomyopathies (both hypertrophic and dilated) and healthy controls. Its unique strength lies in its multi-center (three countries: Spain, Germany, Canada) and multi-vendor (Siemens, GE, Philips, Canon) acquisition protocol. The dataset comprises 150 annotated training images equally distributed across two vendors, and 170 testing cases spread across all four vendors (20 from one vendor, 50 each from three others). Annotations include left ventricle, right ventricle, and left ventricular myocardium at both end-diastolic and end-systolic phases. We utilize the official test set for evaluation and split the training data 95\%/5\% for training and validation.

\noindent\textbf{Radiation Treatment Planning Dataset (StructSeg).} StructSeg~\cite{structseg} comprises specialized CT imaging data focused on radiation therapy planning for nasopharynx and lung cancers. The collection is divided into two primary components: head \& neck (StructSeg H\&N) and thoracic (StructSeg Tho) imaging. The head \& neck portion includes scans from 50 nasopharynx cancer patients with detailed annotations of 22 organs-at-risk (OARs), encompassing crucial structures such as ocular components, brain regions, and maxillofacial structures. The 22 OARs are: left eye, right eye, left lens, right lens, left optical nerve, right optical nerve, optical chiasma, pituitary, brain stem, left temporal lobes, right temporal lobes, spinal cord, left parotid gland, right parotid gland, left inner ear, right inner ear, left middle ear, right middle ear, left temporomandibular joint, right temporomandibular joint, left mandible and right mandible. The thoracic component contains scans from 50 lung cancer patients with annotations of six critical OARs: left lung, right lung,, spinal cord, esophagus, heart, and trachea. We implement a consistent 75\%/5\%/20\%  division for training, validation, and testing across both components.

\begin{figure}[ht]
\begin{center}
\includegraphics[width=\columnwidth]{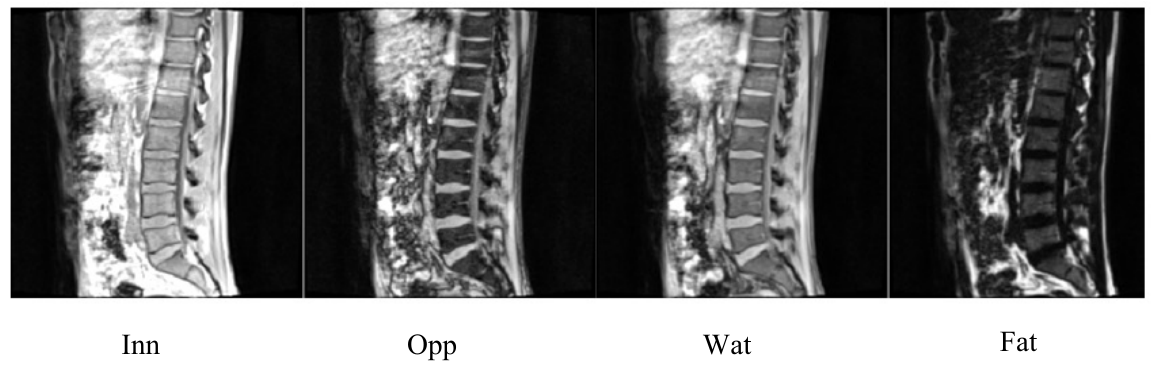}
\end{center}
\vspace{-1em}
\caption{Illustration of four MR modalities of the CSI dataset.}
\label{fig:supp_csi}
\vspace{-1em}
\end{figure}

\noindent\textbf{Spine Imaging dataset (CSI).} CSI~\cite{ivdm3seg} dataset is a specialized collection from the MICCAI Workshop Challenge on Spine Imaging, comprising multi-modal MRI scans of intervertebra discs. The dataset contains 16 complete 3D MRI sets using a Siemens 1.5-Tesla scanner with Dixon protocol, each scan generates four aligned high-resolution 3D volumes (in-phase, opposed-phase, fat, and water images). The imaging focuses on the lower spine, capturing at least 7 intervertebral discs (IVDs) per subject, with expert-annotated binary masks provided for each IVD. We use the four MR modality as separate datasets, namely CSI-inn, CSI-opp, CSI-fat and CSI-wat. The illustration of these four modalities are shown in Figure~\ref{fig:supp_csi}.  We use the CSI-wat in the upstream training, and testing the trained model on CSI-inn, CSI-opp, CSI-fat to evaluate the generalization capability. We can observe that CSI-opp and CSI-inn has relatively similar appearence, where CSI-fat has totally contradictory contrast and intensity, showing great distribution gap.

\noindent\textbf{Automated Cardiac Diagnosis Dataset (ACDC).} The ACDC dataset~\cite{bernard2018deep} consists of cardiac MRI scans collected at the University Hospital of Dijon, covering various cardiac conditions including normal subjects and four pathological groups (myocardial infarction, dilated cardiomyopathy, hypertrophic cardiomyopathy, and right ventricle abnormalities). The scans were acquired using two different Siemens MRI scanners (1.5T and 3.0T) over a six-year period, providing short-axis cardiac images with expert annotations at end-systolic (ES) and end-diastolic (ED) phases. We utilize 100 cases from this collection as a downstream evaluation task to assess our model's generalization capability from the M\&Ms dataset, as they represent different medical centers and scanner configurations while sharing similar anatomical targets.

\noindent\textbf{Thoracic Risk Organ Dataset (SegTHOR).} SegTHOR~\cite{lambert2020segthor} focuses on thoracic organ-at-risk segmentation, providing 40 CT scans with annotations of four critical structures: heart, aorta, trachea, and esophagus. SegTHOR serves as a downstream evaluation task to assess model generalization from StructSeg Tho. We evaluate upstream-trained models directly on all 40 images without additional training.

\noindent\textbf{MSD pancreas \& tumor dataset.}
The MSD pancreas \& tumor dataset is a part of the Medical Image Segmentation Decathlon (MSD)~\cite{antonelli2022medical}, an international challenge aimed at identifying a general-purpose algorithm for medical image segmentation. The competition encompasses ten distinct datasets featuring various target regions, modalities, and challenging attributes. MSD pancreas \& tumor is one of the datasets that is annotated for pancreas and tumors. The shape and position of tumors vary greatly between patients. The MSD pancreas \& tumor dataset consists of 281 CT images. We use it as a downstream task to evaluate models' ability to handle unseen classes, we only use the tumor class for evaluation. We split this dataset into 75\%/5\%/20\% as context/validation/testing set.

\noindent\textbf{Pelvic CT Dataset (Pelvic).} The Pelvic1K dataset~\cite{liu2021deep} is a comprehensive collection of CT scans aggregated from multiple sources, including clinical cases (pre- and post-operative pelvic fractures) and public datasets . These diverse sources provide images with varying field of view, spacing, and clinical conditions, including cases with metal artifacts, vascular sclerosis, and other clinically relevant variations. For our evaluation, we utilize the subset (dataset 6) of Pelvic1K with 103 CT scans with annotations of four skeletal structures: sacrum, left hip bones, right hip bones and lumbar spine. We employ this dataset as a downstream task to assess model performance on novel anatomical structures, using a 75\%/5\%/20\% split for context, validation, and testing respectively.

\begin{table}
  \caption{Datasets statistics. The upper datasets are for upstream training and analysis. The bottom datasets are for downstream tasks on generalization and unseen classes.}
  \label{tab:dataset}
  \centering
   \scriptsize
   \setlength{\tabcolsep}{0.8mm}{
  \begin{tabular}{ccccccc}
    \toprule
    Dataset         & Body Region   & Modality  & Clinical Target   & \#Cls &   Size     \\
    \midrule
    AMOS CT~\cite{ji2022amos}  & Abdomen       & CT        & Organs            & 15        & 300   \\
    AMOS MR~\cite{ji2022amos}  & Abdomen       & MRI       & Organs            & 13        & 60    \\
    AutoPET~\cite{gatidis2022whole} & Whole body & PET & Lesions & 1 & 1014 \\
    BCV~\cite{bcv}     & Abdomen       & CT        & Organs            & 13        & 30    \\
    Brain~\cite{rodrigue2012beta} & Brain & T1 MRI & Structures & 3 & 213 \\
    CHAOS~\cite{CHAOS2021}    & Abdomen       & T1 \& T2 MRI & Organs         & 4         & 60    \\
    KiTS~\cite{heller2019kits19}     & Abdomen       & CT        & Kidney \& Tumor    & 2         & 210   \\
    LiTS~\cite{bilic2019liver}     & Abdomen       & CT        & Liver \& Tumor    & 2         & 131   \\
    M\&Ms~\cite{campello2021multi} & Cardiac & cineMRI & Structures & 3 & 320 \\
    StructSeg H\&N \cite{structseg} & Head \& Neck & CT   & Organs & 22 & 50\\
    StrustSeg Tho\cite{structseg}& Thorax        & CT        & Organs            & 6         & 50    \\
    CSI-wat~\cite{ivdm3seg} & Spine & MR-wat & InterVer Disc & 1 & 16 \\

    \midrule
    ACDC~\cite{bernard2018deep} & Cardiac & cineMRI & Structures &3 & 100\\
    SegTHOR~\cite{lambert2020segthor}  & Thorax        & CT        & Organs            & 3         & 40    \\
    CSI-inn~\cite{ivdm3seg} & Spine & MR-inn & InterVer Disc & 1 & 16\\
    CSI-opp~\cite{ivdm3seg}~\cite{ivdm3seg}  & Spine & MR-opp & InterVer Disc & 1 & 16\\
    CSI-fat~\cite{ivdm3seg} & Spine & MR-fat & InterVer Disc & 1 & 16  \\
    MSD Pancreas~\cite{antonelli2022medical}& Abdomen       & CT        & Pancreas Tumor    & 1         & 281 \\
    Pelvic~\cite{liu2021deep}& Pelvic       & CT        & Bones    & 4         & 103 \\

    \bottomrule
  \end{tabular}}
\end{table}

\section{Supplement Experiments}

\noindent\textbf{Training.}
Iris is trained using an episodic training strategy to simulate in-context learning scenarios. In each training episode, we randomly sample a batch of image-label pairs from our training datasets. For each pair in the batch, we designate it as a reference example and randomly select another pair from the same dataset as the query image. If the sampled data has multiple classes in the mask, we convert it into multiple binary segmentation masks for training.  The training pseudo code is shown in Algorithm~\ref{alg:Iris_training}. 

\begin{algorithm}[h]
\small
\caption{Iris Training}
\label{alg:Iris_training}
\begin{algorithmic}[1]
\State \textbf{Input:} Training dataset $\mathcal{D} = \cup_{k=1}^{K} \mathcal{D}_k$, where $\mathcal{D}_k = \{(\boldsymbol{x}_{k}^i, \boldsymbol{y}_{k}^i)\}_{i=1}^{N_k}$. Image encoder $E$, task encoding module $T$, mask decoder $D$
\While {\textit{not converged}}
    \State // Assemble mini-batch
    \For {b in [1, \dots, batch\_size]}
        \State Sample dataset index $k$ from $[1,K]$
        \State Sample query pair $(\boldsymbol{x}_q, \boldsymbol{y}_q)$ from $\mathcal{D}_k$
        \State Sample reference pair $(\boldsymbol{x}_s, \boldsymbol{y}_s)$ from $\mathcal{D}_k$
    \EndFor
    \State Construct batch $\mathcal{B} = \{(\boldsymbol{x}_q, \boldsymbol{y}_q, \boldsymbol{x}_s, \boldsymbol{y}_s)\}$
    
    \State // Forward pass
    \State Extract task representation $\boldsymbol{T} = T(E(\boldsymbol{x}_s), \boldsymbol{y}_s)$
    \State Predict masks $\boldsymbol{\hat{y}}_q = D(E(\boldsymbol{x}_q), \boldsymbol{T})$
    
    \State // Update
    \State Compute loss $\mathcal{L}_{seg} = \mathcal{L}_\text{dice}(\boldsymbol{\hat{y}}_q, \boldsymbol{y}_q) + \mathcal{L}_\text{ce}(\boldsymbol{\hat{y}}_q, \boldsymbol{y}_q)$
    \State Update parameters of $E$, $D$ and $T$
\EndWhile
\end{algorithmic}
\end{algorithm}

\noindent\textbf{Context Ensemble for Training Classes.}
Previous in-context learning methods require reference image-label pairs even for classes seen during training, leading to two significant limitations. First, the computational overhead of processing reference examples for every inference is unnecessary for previously encountered classes. Second, using only a few context examples often results in suboptimal performance compared to traditional segmentation models, as the task representation may not fully capture the class characteristics learned during training.

\begin{figure*}[ht]
\begin{center}
\includegraphics[width=\textwidth]{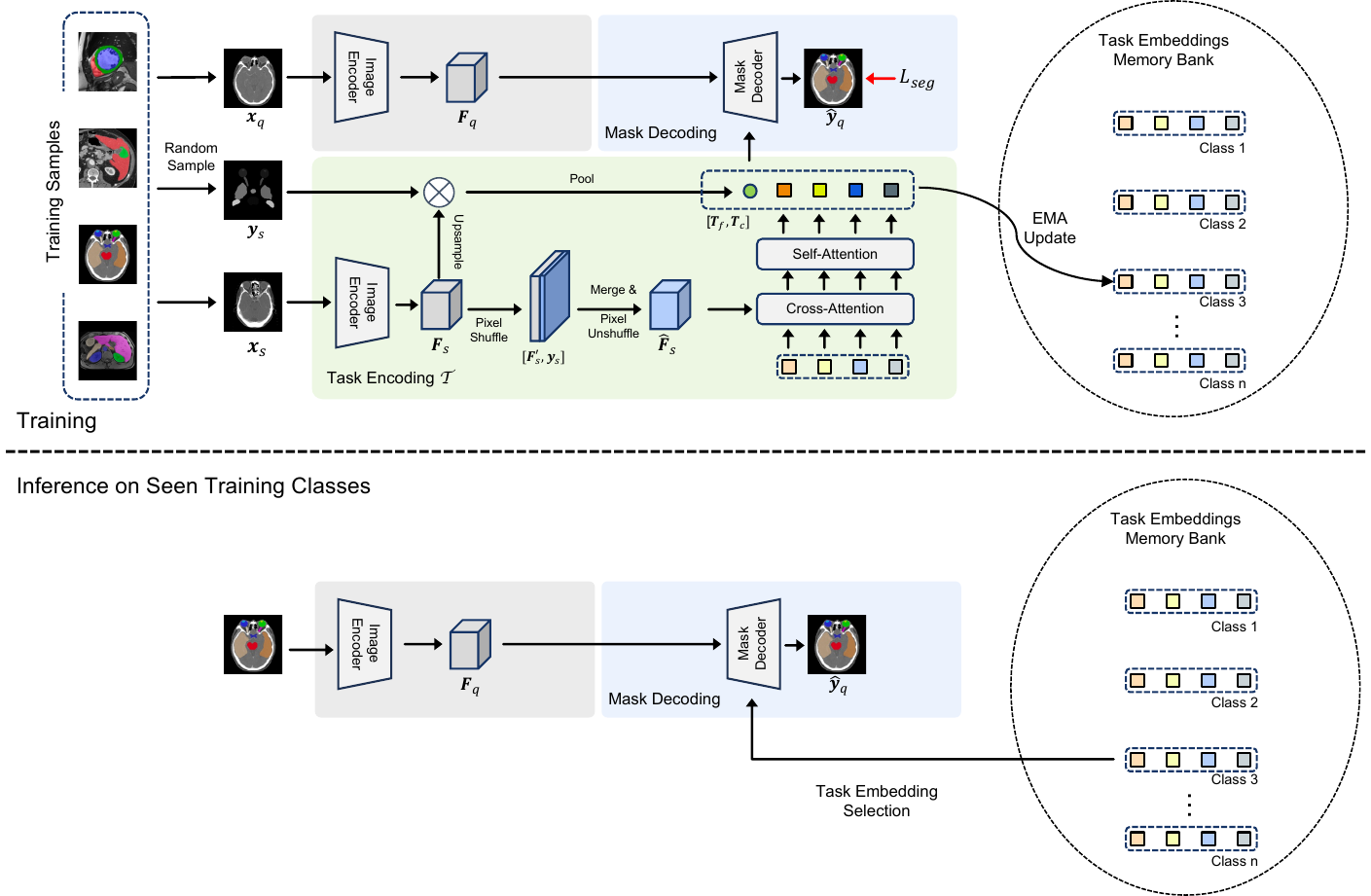}
\end{center}
\vspace{-1em}
\caption{Context ensemble mechanism for efficient handling of training classes. During training, we maintain a memory bank of class-specific task embeddings, updated via exponential moving average (EMA) whenever a class appears in training iterations. At inference, the model directly selects task embeddings from the memory bank for seen classes, eliminating the need for reference examples while maintaining robust performance through accumulated class knowledge.}
\label{fig:supp_context_ensemble}
\vspace{-1em}
\end{figure*}

Instead, we introduce a class-specific task embedding memory bank for classes seen during training that eliminates the need for reference image-label pairs at test time, see Figure~\ref{fig:supp_context_ensemble}. Let $\mathcal{C} = \{c_1, ..., c_K\}$ denote the set of classes seen during training, where $K$ is the total number of training classes. We maintain a memory bank $\mathcal{M} = \{\boldsymbol{T}_1, ..., \boldsymbol{T}_K\}$, where $\boldsymbol{T}_k \in \mathbb{R}^{(m+1) \times C}$ represents the ensemble task embedding for class $k$. During training, when a class $k$ appears in a training iteration, our task encoding module generates a new task embedding $\boldsymbol{T}_k^{new}$ from the reference image-label pair. We then update the corresponding memory bank entry using exponential moving average (EMA):

\begin{equation}
    \boldsymbol{T}_k \leftarrow \alpha\boldsymbol{T}_k + (1-\alpha)\boldsymbol{T}_k^{new}
\end{equation}

where $\alpha=0.999$ is the momentum coefficient. This process gradually accumulates task-specific knowledge across all training samples containing each class, creating robust class representations. During inference on training classes, we can directly select the corresponding task embeddings from $\mathcal{M}$ using class indices from the memory bank, enabling efficient segmentation without the need for reference examples. This mechanism allows Iris to function as both a traditional segmentation model for seen classes and an in-context learner for novel classes.

\textbf{Computation Cost of different inference strategies.} The computational costs of context ensemble and image/object-level retrieval strategies are comparable to the standard Iris implementation. This efficiency stems from our approach of using pre-computed task embeddings, where the overhead for ensemble averaging or similarity-based retrieval is negligible compared to the main inference pipeline. Specifically, retrieval operations add only milliseconds to the total inference time due to their lightweight vector comparison operations. In contrast, in-context tuning requires significantly more computational resources as it involves gradient-based optimization of the task embeddings for each new case, though the tuning process still affects only a small fraction of the model parameters.

\noindent\textbf{Network Architecture.} Our network backbone consists of a 3D UNet with residual connections, comprising four downsampling stages with a base channel dimension of 32. The encoder progressively reduces spatial dimensions while increasing feature channels, and the decoder reconstructs spatial details through skip connections. This architecture effectively captures both local anatomical details and global contextual information in volumetric medical data.

\noindent\textbf{Data Preprocessing.} We implement a standardized preprocessing pipeline to handle the heterogeneous nature of multi-source medical imaging data. First, all volumes are spatially standardized by aligning to a common coordinate system and resampling to an isotropic spacing of $1.5\times 1.5 \times 1.5\ mm$. Intensity normalization is modality-specific: CT images are clipped to the Hounsfield unit range of [-990, 500], while MR and PET images are clipped at their 2nd and 98th percentiles. Finally, z-score normalization is applied to each volume to ensure zero mean and unit standard deviation, facilitating stable network training across different imaging protocols and scanners.

\noindent\textbf{Data Augmentation.} We employ a comprehensive set of augmentation strategies to enhance model robustness. Spatial augmentations include random scaling (0.9 to 1.1), rotation ($\pm$10 degrees), and translation, followed by either random or center cropping to the training size of 128$\times$128$\times$128 voxels. For intensity augmentation, we apply several transformations: multiplicative brightness adjustment (0.9 to 1.1), additive brightness shifts ($\sigma$=0.1), gamma correction (0.8 to 1.2), contrast adjustment (0.8 to 1.2), Gaussian blurring ($\sigma$=0.7 to 1.3), and Gaussian noise ($\sigma\leq$0.02). For reference images, we ensure the preservation of annotated regions after augmentation. These augmentations help simulate various imaging conditions and improve the model's generalization capability across different acquisition protocols and image qualities.

\noindent\textbf{Training and Evaluation Protocol.} During training, Iris processes volumetric data at a window size of $128\times128\times128$ voxels, with random cropping applied as part of our data augmentation strategy to enhance model robustness. For evaluation on large 3D images that exceed the training volume size, we employ a sliding-window inference approach similar to nnUNet~\cite{isensee2021nnu}. This involves moving a $128\times128\times128$ window across the full volume with a 50\% overlap between adjacent windows. Predictions in overlapping regions are averaged to produce smoother segmentation boundaries and reduce edge artifacts. After processing the entire 3D volume, we compute all evaluation metrics (Dice score, etc.) on the complete 3D segmentation result rather than on individual patches, ensuring a comprehensive assessment of the model's performance on anatomical structures of varying sizes and shapes.